\documentclass[conference]{IEEEtran}
\IEEEoverridecommandlockouts

\usepackage{cite}
\usepackage{amsmath,amssymb,amsfonts}
\usepackage{algorithmic}
\usepackage{graphicx}
\usepackage{textcomp}
\usepackage{xcolor}
\def\BibTeX{{\rm B\kern-.05em{\sc i\kern-.025em b}\kern-.08em
    T\kern-.1667em\lower.7ex\hbox{E}\kern-.125emX}}
\begin{document}

\title{PANOPTICON: A PII-Based Assemblage of Naturalistic Output Tokens for Investigating Privacy Leakage Within LLM Context Window
}

\author{\IEEEauthorblockN{Ryan Thornton}
\IEEEauthorblockA{\textit{Department of Computer Science} \\
\textit{Tennessee Tech University}\\
Cookeville, TN, USA \\
rthornton42@tntech.edu}
\and
\IEEEauthorblockN{Mir Mehedi A. Pritom}
\IEEEauthorblockA{\textit{Department of Computer Science} \\
\textit{Tennessee Tech University}\\
Cookeville, TN, USA \\
mpritom@tntech.edu}
\and
\IEEEauthorblockN{Manaak Gupta}
\IEEEauthorblockA{\textit{Department of Computer Science} \\
\textit{Tennessee Tech University}\\
Cookeville, TN, USA \\
mgupta@tntech.edu}
}

\maketitle

\begin{abstract}

Large Language Models (LLMs) are capable of generalizing human language for the completion of never-before-seen tasks, leading to widespread deployment. While this automation provides clear utility,  completing these tasks often requires the insertion of \textit{Personally Identifiable Information} (PII), strings of information that uniquely identify some individual, raising privacy concerns.
However, ethics has prevented the curation of a public, authentic dataset of PII. Without an appropriate dataset, it is difficult to quantify privacy risks.
Thus, we introduce the PANOPTICON pipeline and dataset. The dataset, generated by Meta's {\em Llama-3.1-8B-Instruct} model, contains $67,718$ {\em prompts}, intended for the models context window, containing PII spans derived from 9,674 publicly available synthetic user profiles. We measure lexical diversity and {\em S-BERT} diversity of the created dataset to evaluate realism. Finally, we present a case study showcasing the utility of PANOPTICON data for understanding Prompt Inversion Attacks (PIAs). PANOPTICON thus emerges as the first benchmark dataset for studying PIAs over private corpora, providing a foundation for future LLM privacy research.

\end{abstract}

\begin{figure*}[!t]
    \centering
    \includegraphics[width=0.95\linewidth]{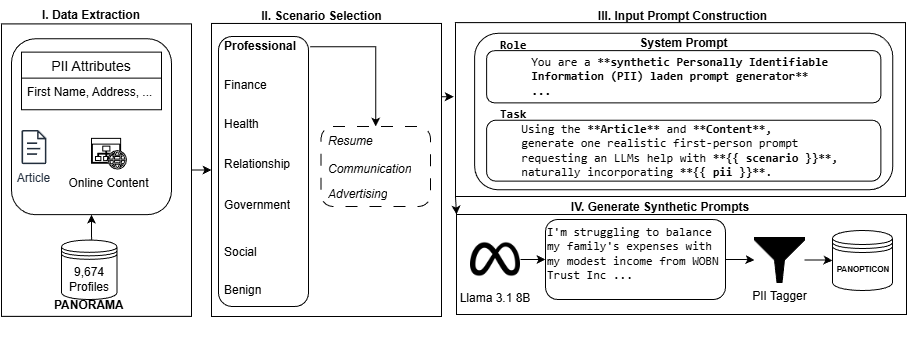}
    \vspace{-0.5em}
    \caption{PANOPTICON Dataset Generation Pipeline}
    \label{fig:pipeline}
\end{figure*}

\section{Introduction}
{LLMs have shown impressive results in complex natural language processing (NLP) tasks \cite{Zhao2023SurveyLLMs}. These models work on tokens passed to their context window, the window that holds all tokens attended over within a particular session. Further, Retrieval Augmented Generation (RAG) and Agentic systems have expanded their utility \cite{singh202agentragsurvey}, leading to widespread adoption. While the full impact of these models, positive or negative, is yet to be understood, they are becoming ubiquitous in multiple domains \cite{chen2024domainsurveylargelanguagemodels}. However, history has taught us that the utility of new technologies is often tempered by new privacy concerns \cite{warren1890right}.

The primary focus of LLM privacy research has been on parametric data, which are data stored in the model, rather than data provided at inference. 
In fact, \cite{mireshghallah2025positionprivacyjustmemorization} claims that 48.4\% of research in the field is on memorization (regurgitating training data). This focus inspired \cite{selvam2025panorama} to release the PANORAMA dataset, a synthetic corpus of Personally Identifiable Information (PII) for studying the leakage of private information due to memorization.
However, privacy researchers have begun exploring the privacy concerns associated with non-parametric data \cite{staab2024beyond, gao2024dory, mireshghallah2024llmssecrettestingprivacy}. Unfortunatley, this field also lacks a corpus of PII for analyzing privacy violations associated with non-parametric data.

To resolve this gap, we introduce \textit{PANOPTICON} (\underline{P}II-based \underline{A}ssemblage of \underline{N}aturalistic \underline{O}ut\underline{P}ut--\underline{T}okens for \underline{I}nvestigating Privacy Leakage Within LLM \underline{CON}text Window), a synthetic dataset designed for controlled study of inference-time PII handling in LLMs. PANOPTICON contains $67,718$ PII-laden prompts spanning six (6) PII categories and eighteen (18) scenarios, generated from $9,674$ synthetic profiles guided by PANORAMA \cite{selvam2025panorama} and produced using Meta's Llama-3.1-8B-Instruct model. To summarize, this paper has the following key contributions:
\begin{itemize}
    \item \textbf{Benchmark dataset generation:} PANOPTICON, a synthetic corpus of 67{,}718 PII-laden prompts spanning six PII categories (healthcare, finance, etc.) and one benign category. The dataset is further split across 38 scenarios (e.g. Budget Planning, Relationship Advice, etc.) with structured metadata to support controlled evaluation. 
    \item \textbf{Dataset characterization:} Quantitative analysis of lexical and semantic diversity, redundancy, and dataset composition to validate PANOPTICON as a broad prompt benchmark for inference-time privacy studies.
    \item \textbf{Case study on benchmark utility:} Presents a case-study on the utility of the dataset with evaluation protocols for use cases like prompt-inversion-style recovery settings that demonstrates how PANOPTICON supports standardized measurements for inference-time PII leakage.
\end{itemize}
The rest of the paper is organized as follows: Section \ref{sec:background} introduces the problem background and motivation for the proposed work. Section \ref{sec:methods} highlights the methodological pipeline for dataset generation, dataset characteristics, content analysis, and use case scenarios for analyzing privacy leakage research in LLMs. Section \ref{sec:discuss} discusses the limitation of the paper, while section \ref{sec:conclusion} concludes the paper and provides future directions. 

\section{Background and Motivation}
In this section, we provide the background research and motivation for this work.

\subsection{Large Language Models (LLMs)} LLMs come in three architectures: Encoder, Decoder, and hybrid models \cite{Roberts_2024}. Regardless of architecture, the model accepts as input a sequence of tokens representing human language. The training objective is to produce the most likely token given the surrounding context. For this paper, we only consider the Decoder model. The training objective of this model is:
$$
P(w_t | w_1,w_2,\ldots,w_{t-1})
$$
In other words, the model predicts the next token given all previous tokens in the sequence. The predicted token is then appended to the sequence and the process repeats until the token limit is reached or until a special end-of-sequence token is produced, making it an \textit{autoregressive} model.

To achieve state-of-the-art results, LLMs must be trained on increasingly large datasets with an increasing number of parameters \cite{Kaplan2020ScalingLF}. Given this requirement, it has become commonplace to utilize multiple web-scrapped datasets containing tens, to hundreds, of GBs worth of publicly available Internet text. For example, ``The Pile'' dataset \cite{pile}, a popular LLM training corpus, contains nearly 900GBs worth of data.

\subsection{Parametric Privacy} Security researchers were quick to recognize the challenges associated with the largeness of the LLMs. Given the large, unstructured nature of the data, and the inexplicability of model output, \textit{memorization} \cite{carlini2019secret} became the primary focus of privacy researchers \cite{carlini2021extracting}. Memorization is the verbatim reproduction of a token, or sequence of tokens, seen in training data, a phenomena particularly common for rare or distinctive sequences \cite{carlini2023quantifyingmemorizationneurallanguage}.

A related attack identified in the literature is the ``Membership Inference Attack" (MIA) \cite{shokri2017membership}. Training-data membership can be reflected in elevated likelihood assigned to member examples in model output, enabling membership and extraction-style privacy attacks. Consequently, sensitive information in the training corpus may be elicited given appropriately crafted inputs~\cite{shokri2017membership,carlini2019secret}. 

Many defenses, such as differential privacy, regular expression-based removal, and homomorphic encryption, have been proposed in literature ~\cite{majmudar2022dpdecoding,subramani2023pi_corpora,giladbachrach2016cryptonets}. However, each of these techniques is coupled with serious drawbacks. The noise introduced by differential privacy results in proportional degradations to model quality \cite{wei2025dualprivpruningefficient}. Regular expressions are overly deterministic, allowing edge cases to persist \cite{rajgarhia2025evaluationstudyhybridmethods}. Homomorphic encryption is prohibitively expensive and thus infeasible at scale \cite{castro2025encryptedllm}. Therefore, implementing these techniques in real-world requires a serious analysis of privacy tradeoffs. 

\subsection{Personally Identifiable Information (PII)} We view PII as any information that can be used to identify a particular individual. In general, PII may be associated with multiple people (e.g. names, zip codes, gender, etc.) or be uniquely associated with some person (e.g. government issued ID numbers.) However, for any given PII category a person possess one identifiable attribute. For example, many people share a first name, but a person only has one first name. It may be objected that some categories, such as allergies, present a valid PII category with multiple values. However, we view this category as containing one unique set. It should be noted that PII has a temporal aspect. While some attributes are practically immutable (e.g. Social Security Number, Blood Type, etc.) others are regularly changed (e.g. Last Name, Address, etc.) Thus, for any given category of PII there exists an ordered set of the attributes whose association has a temporal identifier. For example, if "George" changes his name to "Thomas," we consider this individuals First Name attribute to be ($FirstName_1=George, FirstName_2=Thomas$). However, for simplicity, our work only considers PII for a specific moment in time.

\subsection{PANORAMA Dataset} Understanding the interplay between memorization and PII is only possible if PII exposure can be studied in a controlled setting, where the presence, structure, and semantics of sensitive attributes are explicitly known. In practice, this necessitates datasets with explicit PII annotations (e.g. FirstName: Benjamin.) However, compiling such datasets using real PII associated with real individuals poses significant ethical challenges. To overcome these barriers, the PANORAMA dataset was produced. PANORAMA is a synthetic dataset of user profiles and online content \cite{selvam2025panorama}. By incorporating PANORAMA during training, researchers can better study PII leakage and make more informed decisions regarding privacy tradeoffs.

This dataset contains 9,674 unique profiles with over 30 PII categories (e.g., FirstName, Net worth, Email, etc.) with demographic based attribute distribution. These attributes were fed to an LLM, which then generated, first, a lengthy wiki-style article discussing the synthetic user and, second, a series of fake online content in the voice of the user. These include social media posts, advertisements, product reviews, and more. Each contains some PII such as the users phone number, email, or location. PANORAMA thus serves as an ethical means of researching the dangers associated with parametric memory.

\subsection{Non-Parametric Privacy}
Privacy leakage is not limited to memorization. As demonstrated in \cite{staab2024beyond}, LLMs are capable of inferring sensitive attributes by performing human profiling tasks comparable to, or even exceeding, human professionals. These models were shown to reconstruct latent personal information from seemingly benign inputs. For example, models were able to accurately predict the users location based on word choice.

Further, as demonstrated in \cite{mireshghallah2024llmssecrettestingprivacy}, there is a fundamental mismatch between LLMs and humans when analyzing the contextual relevance of some piece of potentially private information. Thus, as LLMs become more commonplace, privacy leakages are likely to occur at higher rates as individuals, and organizations, put undue trust in systems to protect private information.

Notably, these forms of leakage do not rely on memorization. Instead, they rely on the adversaries ability to reconstruct sensitive information from signals. This has given rise to a new attack vector: Prompt Inversion Attacks (PIAs). In PIAs, adversaries attempt to reconstruct the original input prompts by analyzing model outputs or exposed internal representations \cite{qu2025pia,zhang2024prompt-extraction,gao2024dory}. Further, these prompts often contain PII due to the sensitive domains in which such models operate, making inference-time privacy leakage a significant concern. As with memorization-based leakage, proposed mitigations for PIAs (differential privacy, regex–based filtering, homomorphic encryption, etc.) impose substantial tradeoffs. Thus, meaningful evaluation requires the implementation of a PII-laden dataset focused on non-parametric privacy concerns.
\label{sec:background}

\section{PANOPTICON: Methodology \& Evaluation}
\label{sec:methods}
\subsection{Study Objective}
The PANAROMA dataset \cite{selvam2025panorama} provides a large corpora of PII-laden content, it consists of long, self-contained textual data that a person might post online. However, the length, intent, and relevant PII, are distinct. In contrast, our proposed dataset provides inference time sequences containing PII. This enables controlled studies of inference-time privacy risks, such as Prompt Inversion Attacks, under more realistic conditions.

\begin{figure}[!b]
    \centering
    \includegraphics[width=0.80\linewidth]{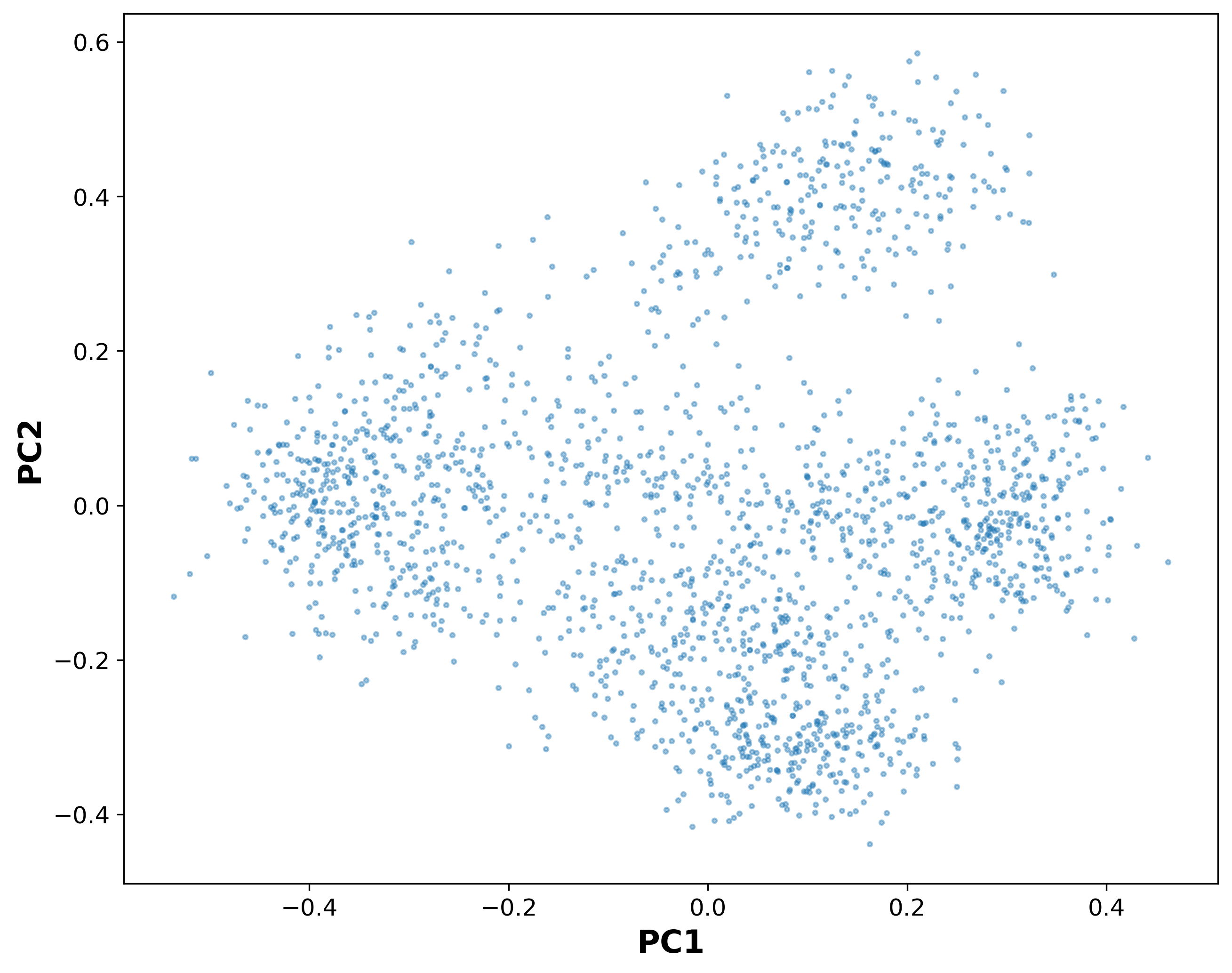}
    \caption{PCA projection of Sentence-BERT embeddings for all PANOPTICON prompts (each point corresponds to a single prompt)}
    \label{fig:sbert_pca}
\end{figure}

\begin{figure*}[!t]
    \centering
    \includegraphics[width=0.99\linewidth]{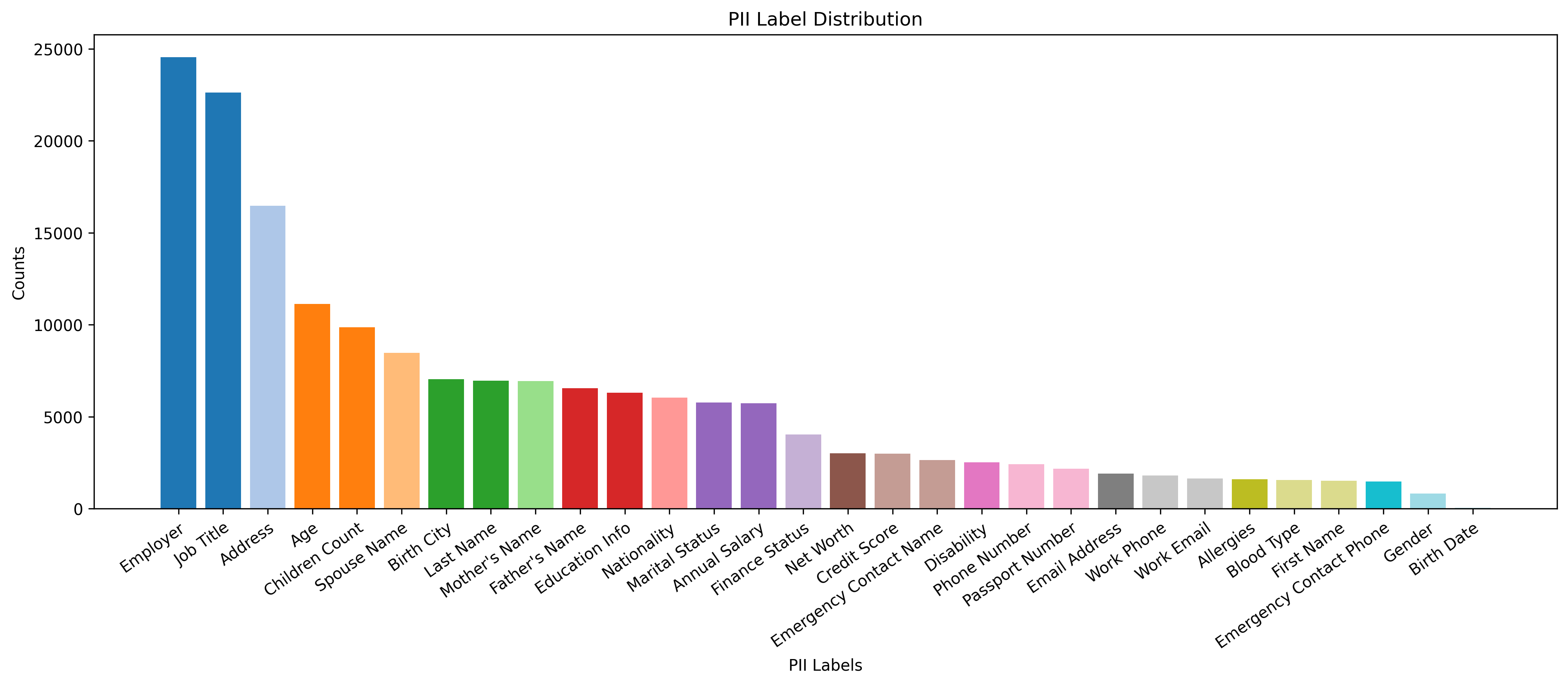}
    \caption{PII type distribution across PANOPTICON dataset prompts}
    \label{fig:pii_dist}
\end{figure*}

\subsection{PANOPTICON Dataset Generation Pipeline}
As shown in Figure \ref{fig:pipeline}, our proposed PANOPTICON dataset generation pipeline is a 4-step process. We iterate through the PANAROMA dataset extracting user data, selecting scenarios, constructing an input, and generating a series of synthetic prompts containing PII. For our purposes, a `prompt' is a sequence of English text written with the intent of being passed to an LLM. The details of the process is described below:

\noindent \textbf{Step-I (Data Extraction):} This phase selects data from PANORAMA. Specifically, we extract the wiki-style article, a random sample of online posts, and the list of PII. The wiki article contextualizes the individual. The online content is used to inform stylistic decisions. Finally, the list of PII serves as a pool of insertion candidates.

\noindent \textbf{Step-II (Scenario Selection):} This phase establishes the context. This context helps reduce the pool of candidate PII to a more realistic subset. First, we iterate through seven (7) pre-selected PII categories, {\em Professional}, {\em Finance}, {\em Health}, {\em Relationship}, {\em Government}, {\em Social Media} and  {\em Benign} (The {\em Benign} category has twenty (20) scenarios to ensure diversity of benign prompts that ideally do not contain any PII). For each category, there are is a list of potential scenarios (twenty for benign, 3 for all others). Each profile generates a single prompt for each category with the scenario being randomly selected. This step serves to diversify potential PII in the proper category. For example, ``Blood Type'' is only considered as a candidate in the {\em Health} category.
    
\noindent \textbf{Step-III (Input Construction):} This phase utilizes the retrieved context (profile information and context), organizing it for model input. After the profile has been selected and the scenario determined, the model feed the article, online content, scenario, and user PII into a custom system prompt. This organizes all data thus far collected and explains how to use it for generation (further details in \textit{Appendix A}).

\noindent \textbf{Step-IV (Generate Synthetic Prompts):} The final phases passes the System Prompt to the chosen LLM (i.e, {\em Llama-3.1 8B} in this case study). This produces the final synthetic prompt with selected PII. For elaboration on how ground-truth labels are generated, see below. 

Note that, Step-I repeats for $N$ times. Steps II-IV are repeated until all 7 categories are exhausted. The process generates a total of $(N \times 7 \times 1)$ synthetic prompts, where $N=9,674$ corresponding to all the unique profiles in {\em PANAROMA} dataset. 
The prompts used for initialization, formatting, and constraints can be found in Appendix A and B. Further, the specific scenarios used, and the distribution of those scenarios, across the 7 categories is detailed in Table III within the Appendix.
 
\subsection{Dataset Composition Overview}

Our PANOPTICON dataset consists a total of $67,718$ rows and six columns. The six columns are \textit{ID}-- representing unique ID for synthetic profiles, \textit{Category}-- representing one of seven potential PII categories, \textit{Scenario}-- representing one of 38 potential scenarios (i.e., 20 benign scenarios, and 18 private), \textit{Prompt}-- representing the generated prompt, \textit{PromptPII}-- representing the PII labels, and \textit{Content}-- representing the sampled online content used to guide generation.

We measure the lexical diversity of the dataset using token-level statistics computed over the entire dataset. PANOPTICON contains a total of $4,073,192$ tokens spanning $65,783$ unique token types, yielding a type–token ratio (TTR) of $0.016$. While TTR naturally decreases as corpus size increases, the observed value is consistent with large, naturalistic text corpora and indicates substantial vocabulary breadth. To further characterize lexical variability, we compute token entropy, obtaining a value of $9.19$ bits. High token entropy suggests that token usage is broadly distributed rather than dominated by a small subset of frequent tokens, supporting the claim that PANOPTICON does not collapse into a narrow or template-driven language distribution.

Because lexical measures alone do not capture semantic variation, we also evaluate semantic diversity using Sentence-BERT embeddings. We compute the average pairwise cosine distance across all generated prompts, obtaining a value of $0.83$. This indicates that, on average, prompts are semantically well-separated rather than clustered around a small set of canonical meanings. This can be seen visually in Figure \ref{fig:sbert_pca}. This figure illustrate the Sentence-BERT embedding diversity, showing four distinct groupings of lexical similarity. This supports claims of diversity without indicating degeneration. To quantify redundancy directly, we measure the fraction of prompt pairs with cosine distance below $0.05$, which we treat as near-duplicates. The resulting redundancy rate of $5.36\%$ suggests that only a small minority of prompts exhibit strong semantic overlap. This overlap is primarily concentrated in the benign category, as it has the least potential for diverse inspiration.

Together, these lexical and semantic metrics indicate that PANOPTICON exhibits high diversity at both the surface and meaning levels. This diversity is critical for evaluating inference-time privacy attacks, as prompt inversion behavior may vary substantially with linguistic structure, intent, and semantic context. By validating that PANOPTICON avoids excessive repetition or semantic collapse, these results support its use as a realistic and robust benchmark for studying PII leakage under prompt inversion attacks.

Figure \ref{fig:pii_dist} denotes the distribution of PII labels across generated prompts. The graph indicates that certain PII attributes, such as `Employer' and `Job Title' are more common, while others such as `Gender' or `Birth Date' are rarely included. Intuitively, certain PII will be more common than others. We find it likely that productivity related PII would be more common. However, measuring the real distribution is impossible without access to real prompts. In any case, PII of certain types will need to be more or less predominate given the field studied. Thus, our pipeline, with minimal tweaks, would enable changing this distribution to fit the scenario.

\begin{table}[!t]
\centering
\caption{PANOPTICON Dataset and PII Content Distribution}
\label{tab:pii_leakage}
\resizebox{0.85\linewidth}{!}{%
\begin{tabular}{l l l r r r r r}
\hline
\textbf{Category} & \textbf{Contains PII} & \textbf{Lacks PII} & \textbf{Total} & \textbf{\% Contains PII} \\
\hline
Professional & 9,632 & 42 & 9,674 & 99.57 \\
Finance & 9,616 & 58 & 9,674 & 99.40 \\
Health  & 9,252 & 422 & 9,674  & 95.64 \\
Relationship & 9,253 & 421 & 9,674 & 93.44 \\
Government & 9,039 & 635 & 9,674 & 93.44 \\
Social & 9,188 & 486 & 9,674 & 94.98 \\
Benign & 818 & 8,856 & 9,674 & 8.46 \\
\hline
ALL & 56,798 & 10,920 & 67,718 & 83.87 \\
\hline
\hline
\end{tabular}
}
\end{table}

Additionally, table \ref{tab:pii_leakage} shows prompt distributions for prompts with, and without, PII. We observe that prompts expected to contain PII occasionally lack it, while benign prompts, ideally free of PII, on rare occasion contain PII. Moreover, the complex nature of PII, and what constitutes true inclusion, is challenging to ascertain. For example, assume a person lives at ``308 Negra Arroyo Lane''. If the number ``308'' or the word ``Lane'' is present, whether this constitutes meaningful PII leakage or not is debatable.

However, to generate ground-truth labels we must have some reporting mechanism. Relying on the LLM to self-report PII inclusion would lead to uncertainty, while a deterministic PII tagging mechanism suffers issues already discussed. Still, we decided to utilize a regular expression based approach. After the prompt is generated, a function checks the prompts for PII utilizing our defined rules.

Unfortunately, the deterministic nature of our PII tagging mechanism may have overly captured PII in some instances, while under capturing PII in others. Further, it is possible that the model fails to include (or omit) PII. Despite these limitations, all PII categories have, at least, a $93.44\%$ PII presence while the benign category only contains $8.45\%$ prompts with PII. This indicates broad compliance with PII constraints, and thus still provides meaningful ground-truth for future experiments.

\section{Experimental Case Study and Results}
\label{sec:experiment}
To test the potential utility of our PANOPTICON dataset, we conduct a limited study of prompt inversion against the PANOPTICON data. Specifcally, we adapt the threat model of Qu et al. \cite{qu2025pia}. However, our goal is not the demonstration of state of the art prompt inversion, but to provide an example of PANOPTICON's utility.

\subsection{Threat Model}
In this study, we implement a Prompt Inversion Attack (PIA) in a collaborative inferencing environment. Here, an LLM is partitioned across multiple participants. Each participant provides resources by handling some of intermediate activations. The threat actor, who is ``passive but curious,'' has white box access to the system.

Specifically, an adversary attempts to reconstruct the original prompt text via an inversion procedure that (i) performs constrained optimization in embedding space to recover a sequence of token embeddings whose forward activations match the observed boundary activations, and (ii) applies an {\em adaptive discretization} stage that maps optimized embeddings back to discrete tokens using activation calibration and an LLM-driven semantic candidate set. The original work evaluates attack success using 3 primary metrics {\em token-accuracy} (i.e., exact positional recovery), {\em BLEU} (i.e., overlap-based similarity), and {\em NERR} (i.e., named-entity recovered ratio), thereby capturing complementary notions of strict reconstruction, surface-level overlap, and entity-level fidelity.

\subsection{Experimental Setup}
We adopt a similar high-level evaluation criteria to Qu et al.~\cite{qu2025pia}: 
collaborative inference with the attacker modeled as the final participant (i.e., the most challenging placement), which maximizes the number of preceding layers that must be inverted. However, we introduce three changes in our experimental setup aligned with our research goals and with practical resource constraints:

\begin{enumerate}
    \item \textbf{Target model scale and tuning.} Qu et al.~\cite{qu2025pia} report results primarily using a 65B-parameter base model (LLaMA-65B / LLaMA-1). Due to the computational cost of replicating 65B-scale white-box inversion at our evaluation scale, we instead target Meta's \textit{Llama-3.1-8B-Instruct}. This substitution changes both model size and training objective (instruction tuning), and thus may materially alter the geometry of internal representations and the ease of inversion.
    \item \textbf{Dataset domain and objective.} Whereas Qu et al.~\cite{qu2025pia} primarily evaluate on the benign Skytrax corpus (airline reviews), we evaluate on our PANOPTICON data, which is explicitly constructed to include privacy-relevant prompt structures and labeled PII. Our goal is not to reproduce state-of-the-art inversion accuracy but rather to obtain a consistent, lightweight leakage proxy for dataset evaluation. 
    \item \textbf{Relevant Metrics} We continue to use Token Level Accuracy. 
    However, we avoid BLEU/NERR-style string similarity metrics because they are sensitive to paraphrase, formatting, and non-PII boilerplate, and therefore conflate general textual similarity with privacy-relevant recovery; instead, PANOPTICON’s span annotations enable a leakage-focused PII-F1 that explicitly captures the precision–recall tradeoff in PII extraction.
\end{enumerate}

With the above alterations taken into consideration, from this experiment, we-- (i) capture boundary hidden states from the final transformer block to emulate strong attacker placement, (ii) produce an initial discrete sequence via a nearest-neighbor embedding inversion baseline, and (iii) optionally apply a lightweight local refinement via greedy coordinate search. We also have two variants of our experiment settings with our target model and the PANOPTICON dataset. \textit{First}, with refinement disabled (i.e., OFF), we accept the nearest-neighbor token at each position as the final prediction. \textit{Second}, with refinement enabled (i.e., ON), we perform a bounded local search over candidate replacements and select tokens that improve similarity between the resulting hidden state and the recorded target activation. Refinement increases computational cost, and thus we opted to utilize a smaller sample size. In our implementation we use cosine distance and {\em k} embedding neighbors plus {\em m} candidates per position, for 1–3 refinement passes; we keep these hyper-parameters fixed across all PANOPTICON data slices. 
 Appendix C presents more details on refinement toggle and different embedding-based projections.  

\begin{table}[!t]
\centering
\small
\caption{State-Of-The-Art White-Box Prompt Inversion Attack (PIA) 
Vs. Our PIA on PANOPTICON Dataset 
}
\resizebox{1\linewidth}{!}{
\begin{tabular}{l l l r r r r r}
\hline
\textbf{Exp. Setting} & \textbf{Target Model} & \textbf{Dataset} & \textbf{\# prompts} & \textbf{Tok-Acc} & \textbf{PII-F1} \\
\hline
Qu et al. (white-box PIA) & LLaMA-65B (base) & Skytrax & 150 & 0.8838 & -- \\
Ours (approx-PIA, refine OFF) & Llama-3.1-8B-Instruct & PANOPTICON (PII-only) & 500 & 0.0093 & 0.0242 \\
Ours (approx-PIA, refine ON)  & Llama-3.1-8B-Instruct & PANOPTICON (PII-only) & 50  & 0.0211 & 0.0044 \\
\hline
\end{tabular}
}
\label{tab:pia_compare}
\end{table}

\subsection{Experiment Results}
Table~\ref{tab:pia_compare} summarizes the reported white-box PIA performance from Qu et al.~\cite{qu2025pia} under the 4-participant setup, last-participant attacker placement, alongside our inversion results. It seems, the original PIA achieves high inversion fidelity (token accuracy $\approx 0.88$), indicating that boundary activations can retain substantial recoverable information in collaborative inference deployments. In contrast, our lightweight approximation yields substantially lower absolute reconstruction quality on PANOPTICON. This degradation is expected given: (i) the smaller, instruction-tuned target model, (ii) the domain shift from benign reviews to explicitly privacy-relevant prompts, and (iii) the fact that our approximation omits the full constrained optimization and full-vocabulary adaptive discretization. 

Nonetheless, these results should be interpreted in two layers. \textit{First}, in absolute terms, our PIA-inspired approximation yields substantially lower reconstruction fidelity than the full white-box Prompt Inversion Attack (PIA) reported by Qu et al.~\cite{qu2025pia}. \textit{Second}, despite low absolute fidelity, the proxy remains suitable for our study objective: it produces consistent, PII-aware recovery measurements over a large, labeled prompt corpus. In particular, because PANOPTICON provides explicit PII span labels and structured prompt metadata, we can test whether recovery differs systematically between PII spans and surrounding benign context, and whether prompts containing PII exhibit different inversion behavior than benign prompts. Finally, the observed PII reconstruction rates under this attacker placement provide an empirical lower bound on PII recoverability under our resource-constrained evaluation setting.
Further information on token choice and Nearest Neighbore / Cosine Similarity can be found in Appendix D.

\section{Discussion and Limitation}
\label{sec:discuss}
PANOPTICON dataset is designed to evaluate the privacy risks of non-parametric data through the lens of PII. This is accomplished via a dataset of diverse, synthetically generated, prompts containing explicit PII spans and ground truth labels. This structure enables controlled, label-aware comparisons between PII spans and surrounding benign context, thereby operationalizing a consistent leakage probe for evaluation. As a result, PANOPTICON supports systematic measurement of recoverability across prompt types, PII categories, and model settings, strengthening empirical inquiry into PII-specific behavior and the measurability of such differences. Further, Using a PIA-inspired inversion probe, we quantify how much labeled PII can be recovered from boundary activations and report PII-centric recovery statistics at scale. These measurements provide a practical baseline for PII recoverability under inversion-style leakage and establish a repeatable protocol for future evaluations with stronger attackers or additional targets.

\subsection{Limitations}
\label{sec:limitations}
While this is the first benchmark dataset and we show it's practical utility through case study, this paper has the following limitations. \textit{First}, the complex nature of PII serves to undermine our PII tagging strategy. Unfortunately, it is the case that the tagger either claimed PII existed within a prompt, when it did not, or the model inserted PII into the prompt, despite being told not to. In either case, this diminishes ground-truth accuracy. \textit{Second}, computational limitations impose restrictions on model choice. However, we hope our results and model choice demonstrate the flexibility of our pipeline for various models.
\textit{Third}, lack of real-world data presents problems for establishing the realism of PII utilization. The rate users provide certain PII spans is unknown, and thus our presentation of PII distributions are, at best, rough reflections of the real world. However, this can be rectified in particular cases where a different distribution is required.

\section{Conclusion and Future Directions}
\label{sec:conclusion}

In summary, this paper introduces PANOPTICON, a large-scale dataset of A PII-based Assemblage of Naturalistic Output Tokens for Investigating Privacy Leakage in Conversational AI with explicit span-level PII labels and structured metadata, designed to enable controlled investigation of privacy leakage in conversational LLM settings. Motivated by the implicit interchangeability assumption induced by benign evaluation corpora, we formalized research questions targeting whether PII behaves differently from benign tokens and whether such differences can be measured systematically. We further provided a case study using a computationally feasible, PIA-inspired inversion probe to obtain PII-centric recovery statistics over labeled prompts. While this proxy does not reproduce state-of-the-art white-box inversion performance, it provides a consistent, repeatable leakage measurement framework that can be applied across PANOPTICON slices and future model targets. We release PANOPTICON and the accompanying generation and evaluation pipeline to support standardized benchmarking, facilitate stronger attacker evaluations in future work, and advance empirical understanding of inference-time privacy behavior for PII-bearing prompts.

Future research can extend our work to a broader suite of prompt inversion attacks which would strengthen external validity by characterizing how leakage trends on PANOPTICON vary across attacker families and threat models, and by enabling a more granular analysis of recoverability across PII categories (e.g., addresses vs. names vs. phone numbers). Greater computational resources would support larger models, improving scale and diversity.

\section*{Acknowledgment}
This work is partially supported by the NSF Research Traineeship (NRT) grant \# 2346001, \# 2416990 and  National Security Agency award H98230-24-1-0102 at Tennessee Tech University.

\bibliographystyle{IEEEtran}
\bibliography{sections/ref}

\appendix
\section{System Prompt Construction Process}
\subsection{Initialization Prompts}
We provide the following role-playing instruction to LLM: ``You are a synthetic Personally Identifiable Information (PII) laden prompt generator. Your task is to use the provided synthetic input data to generate a realistic, first-person prompt that can be used for PII research in Large Language Model (LLM) prompt studies.''

Next, we set the objective through the following prompt: ``Generate a single synthetic user prompt written in the first-person voice of the synthetic profile. The prompt must be a natural request for help with {\tt <scenario>}, which falls under the broader category of {\tt <category>}. The prompt must naturally include PII elements woven into the text as part of the scenario (never in list form). Here is a suggested pool of PII to use: {\tt <pii-list>}.''

Then, we provide additional input instructions as follows:\\
``You will be given:\\
\noindent - Article: A long, wiki-style description of the synthetic person. Use it only for general background; do not copy sentences.\\
\noindent- Content: Short writing samples from the synthetic person. Use these as tone, phrasing, and style cues.''

For output formatting we provide the following style prompt:\\
``Your response must contain only one thing: A single, coherent, first-person prompt requesting help from an LLM with {{ scenario }}, written in the style of the synthetic person, and naturally containing at least some of this PII: {{ pii }}. Limit the prompt to 2-3 sentences. Do not include explanations, headers, bullet points, or any surrounding text. Output just the prompt itself.''

Lastly, we provide the following prompt for generating the final naturalistic output prompt to include in the PANOPTICON dataset: ``Using the Article and Content, generate one realistic first-person prompt requesting an LLMs help with {\tt <scenario>}, naturally incorporating pii {\tt <pii>}.''

\subsection{Mandatory Constraints}
We provide the following mandatory constraints:
\begin{itemize}
    \item ``The prompt must be written in first person, as if typed directly by the synthetic person.  
    \item It must clearly be a request for assistance (e.g., drafting, revising, preparing, summarizing), not a biography or third-person description.  
    \item It must pertain specifically to the area of {{ category }}, and more precisely {{ scenario }}.
    \item The PII values in {{ pii }} must be smoothly integrated into natural sentences, not listed or formatted as key:value pairs. The goal is not maximal PII, but natural PII. 
    \item Do not begin with phrases such as “As I…”, “As I approach…”, “As I get ready…", "I am seeking, etc. be creative with prompt framing based on how you percieve the individual based on the content.
    \item Keep the writing concise, coherent, and consistent with the tone of “Content.”  
    \item Do not include any text other than the single generated prompt
    \item The final output should be a prompt directed towards an LLM''
\end{itemize}

\section{PANOPTICON Prompt Distribution Across Scenarios}
\begin{table}[!t]
\centering
\caption{Overall Distribution of PII Usage Categories, Corresponding Scenarios, and Prompt Counts}
\resizebox{0.80\linewidth}{!}{
\begin{tabular}{l l l}
\hline
\textbf{Category} & \textbf{Scenarios} & \textbf{\# Prompts} \\
\hline
 & Resume, CV or Career Materials & 3,208\\
Professional & HR and Internal communications & 3,288\\
 & Advertise Professional Services & 3,178 \\
\hline
 & Budget planning & 3,201\\
Finance & Tax bracket/deduction inquiry & 3,220 \\
 & Mortgage or loan application & 3,253 \\
\hline
 & Medical Provider Communications & 3,288\\
Health & Medical history summary & 3,327\\
 & Insurance, Billing, or Administrative Assistance & 3,059\\
\hline
 & Message to partner/family member & 3,232 \\
Relationship & Relationship Advice & 3,230 \\
 & Family trip planning & 3,212\\
\hline
 & Document renewal or recovery assistance & 3,178\\
Government & Social services inquiry & 3,229 \\
 & Identity verification request & 3,267 \\
\hline
 & Social media post & 3,200\\
Social & Email or Direct Message assistance & 3,215\\
 & Online Marketplace Assistance & 3,259\\
\hline
 & Productivity tips & 467\\
 & Travel recommendations & 479\\
 & Book/movie suggestions & 527\\
 & General knowledge questions & 462\\
 & Cooking recipes & 467\\
 & Gardening advice & 493\\
 & Fitness routines & 488\\
 & Language learning assistance & 476 \\
 & Creative writing prompts & 432\\
Benign & DIY project ideas & 478\\
 & Meditation and mindfulness techniques & 506\\
 & Time management strategies & 489\\
 & Hobby exploration suggestions & 498\\
 & Study tips and techniques & 475\\
 & Home organization hacks & 469\\
 & Stress management methods & 468\\
 & Pet care advice & 501\\
 & Outdoor activity recommendations & 498 \\
 & Video game suggestions & 510\\
 & Music or podcast recommendations & 492\\
\hline
\end{tabular}
}
\label{tab:scenario_categories}
\end{table}

\section{Refinement Toggle and Embedding-Based Token Projection}
\label{app:refinement_tokenization}

This appendix provides additional implementation detail for (i) the \emph{refinement} toggle used in our approximate prompt inversion pipeline (``refine off'' vs.\ ``refine on'') and (ii) the embedding-based token projection strategy used to convert continuous representations into discrete vocabulary tokens via nearest-neighbor retrieval under cosine similarity.

\subsection{Refinement Process (On vs.\ Off)}
\label{app:refinement_on_off}

Our inversion pipeline produces an \emph{initial} discrete prompt hypothesis by projecting a continuous representation (e.g., an optimized embedding sequence or intermediate latent token representation) onto the model vocabulary. We then optionally apply a lightweight \emph{discrete refinement} stage to improve syntactic plausibility and local coherence. We report results for both settings:

\paragraph{Refine Off.}
When refinement is disabled, we perform a single-pass projection from the continuous representation to a discrete token sequence:
\begin{enumerate}
    \item Obtain a continuous token-level representation $\{\mathbf{z}_i\}_{i=1}^{L}$ (length $L$).
    \item Project each $\mathbf{z}_i$ to a discrete token via nearest-neighbor lookup in embedding space (Appendix~\ref{app:token_projection}).
    \item Detokenize the resulting token sequence into text and report metrics.
\end{enumerate}
This setting isolates the \emph{projection step} and avoids any additional search over the discrete space.

\paragraph{Refine On.}
When refinement is enabled, we apply an additional local search procedure over the discrete sequence to mitigate common failure modes of one-shot projection (e.g., malformed punctuation, fragmented BPE segments, or locally inconsistent phrasing). Concretely, refinement iterates over token positions and proposes small edits drawn from an embedding-similarity candidate set:
\begin{enumerate}
    \item Initialize with the projected token sequence $\hat{\mathbf{t}} = (\hat{t}_1,\ldots,\hat{t}_L)$ from the ``refine off'' procedure.
    \item For each position $i$, retrieve a small candidate list $\mathcal{C}_i$ (top-$K$ tokens) using cosine similarity between $\mathbf{z}_i$ and vocabulary embeddings (Appendix~\ref{app:token_projection}).
    \item Score candidate replacements under a local objective. In practice, we use a context-conditional language-model score (negative log-likelihood / perplexity proxy) computed for the reconstructed sequence in a limited window around position $i$.
    \item Accept a replacement $\hat{t}_i \leftarrow c$ if it improves the objective by a margin $\Delta$; otherwise keep $\hat{t}_i$ unchanged.
    \item Repeat for up to $R$ refinement rounds or until no positions accept updates (early stopping).
\end{enumerate}

\noindent\textbf{Rationale.} ``Refine on'' adds a small amount of discrete exploration while keeping computation bounded (controlled by $K$ candidates per position and $R$ rounds). It is particularly helpful for PII-heavy strings (emails, phone numbers, IDs), where tokenization can be brittle and a single projection pass frequently produces near-miss fragments.

\subsection{Token Choices via Nearest Neighbor / Cosine Similarity}
\label{app:token_projection}

A key design decision is how to map continuous token-level representations $\mathbf{z}_i \in \mathbb{R}^{d}$ back into discrete tokens from a fixed vocabulary $V$. We adopt a nearest-neighbor projection in the model's input embedding space, using cosine similarity as the distance metric.

\paragraph{Vocabulary embedding matrix.}
Let $E \in \mathbb{R}^{|V| \times d}$ denote the model's input embedding matrix, where the row vector $\mathbf{e}_v$ is the embedding for token $v \in V$. We precompute $\ell_2$-normalized embeddings:
\[
\tilde{\mathbf{e}}_v = \frac{\mathbf{e}_v}{\|\mathbf{e}_v\|_2}, 
\qquad
\tilde{\mathbf{z}}_i = \frac{\mathbf{z}_i}{\|\mathbf{z}_i\|_2}.
\]

\paragraph{Cosine-similarity nearest neighbor.}
For each position $i$, we select the token maximizing cosine similarity:
\[
\hat{t}_i = \arg\max_{v \in V} \ \cos(\mathbf{z}_i, \mathbf{e}_v)
= \arg\max_{v \in V} \ \tilde{\mathbf{z}}_i^\top \tilde{\mathbf{e}}_v.
\]
This is equivalent to nearest-neighbor search on the unit hypersphere. In practice, we retrieve the top-$K$ tokens
\[
\mathcal{C}_i = \text{TopK}_{v \in V}\big(\tilde{\mathbf{z}}_i^\top \tilde{\mathbf{e}}_v\big),
\]
which supports both (i) one-shot decoding (choose the top-1 token) and (ii) refinement (evaluate multiple candidates).

\paragraph{Why embedding-based retrieval?}
We favor embedding-space projection for three reasons:
\begin{enumerate}
    \item \textbf{Tokenization robustness:} BPE tokenization can split PII into multiple sub-tokens (e.g., digits, punctuation, domain fragments). Embedding-space proximity often yields plausible alternative sub-tokens when exact reconstruction is difficult.
    \item \textbf{Candidate diversity:} The top-$K$ nearest neighbors provide a compact, semantically related candidate set $\mathcal{C}_i$ without relying solely on a single greedy choice.
    \item \textbf{Efficiency:} With normalized embeddings, cosine retrieval reduces to dot products and can be accelerated via standard approximate nearest-neighbor techniques if needed.
\end{enumerate}

\paragraph{Practical candidate constraints (PII-aware filtering).}
Because PANOPTICON contains prompts with explicit PII spans, we optionally apply lightweight filters when constructing $\mathcal{C}_i$:
\begin{itemize}
    \item \textbf{Character-class constraints:} for positions within numeric-heavy spans, prioritize digit/punctuation tokens (e.g., tokens containing \texttt{0--9}, \texttt{-}, \texttt{/}, \texttt{.}, \texttt{@}) and deprioritize purely alphabetical tokens.
    \item \textbf{Control-token exclusion:} exclude special/control tokens not meant to appear in natural text.
    \item \textbf{Whitespace/prefix handling:} handle leading-space tokens (common in BPE) consistently so that reconstructed text does not accumulate spurious spacing artifacts.
\end{itemize}

\paragraph{Refinement scoring objective (local LM score).}
During refinement, each candidate replacement $c \in \mathcal{C}_i$ is evaluated by re-scoring the sequence under the model in a bounded context window. Let $\hat{\mathbf{t}}^{(i \leftarrow c)}$ denote the sequence with token $i$ replaced by $c$. We compute a local negative log-likelihood proxy:
\[
\mathcal{L}( \hat{\mathbf{t}}^{(i \leftarrow c)} ) 
= - \sum_{j \in \mathcal{W}(i)} \log p_\theta\!\big(\hat{t}_j \mid \hat{\mathbf{t}}_{<j}\big),
\]
where $\mathcal{W}(i)$ is a small index window around $i$ (to keep computation tractable). We accept the replacement if it reduces $\mathcal{L}$ by at least a margin $\Delta$.

\end{document}